\documentclass{article}
\usepackage{spconf,amsmath,graphicx}
\usepackage{subfigure}
\usepackage{booktabs}
\usepackage{makecell}
\usepackage{multirow}
\usepackage{appendix}
\usepackage{bbding}

\title{AN ENSEMBLE MODEL FOR DISTORTED IMAGES IN REAL SCENARIOS}
\name{Boyuan Ji, Jianchang Huang, Wenzhuo Huang, Shuke He}
\address{Zhejiang University of Science and Technology\\
	Edge Intelligence Security Lab, School of Big Data Science\\
	Hangzhou, Zhejiang, China}
\begin{document}
%
\maketitle
\begin{abstract}
Image acquisition conditions and environments can significantly affect high-level tasks in computer vision, and the performance of most computer vision algorithms will be limited when trained on distortion-free datasets. Even with updates in hardware such as sensors and deep learning methods, it will still not work in the face of variable conditions in real-world applications. In this paper, we apply the object detector YOLOv7 to detect distorted images from the dataset CDCOCO. Through carefully designed optimizations including data enhancement, detection box ensemble, denoiser ensemble, super-resolution models, and transfer learning, our model achieves excellent performance on the CDCOCO test set. Our denoising detection model can denoise and repair distorted images, making the model useful in a variety of real-world scenarios and environments.
\end{abstract}
\section{Introduction}

CNN-based methods have become prevailed in object detection~\cite{1,2} and image classification~\cite{17,18,19}. They not only have achieved promising performance on benchmark datasets~\cite{3,4} but also have been deployed in the real-world applications such as autonomous driving~\cite{5} Due to the domain shift in input images~\cite{6}, general object detection models trained by high-quality images often fail to achieve satisfactory results in detecting distorted images (e.g., foggy and blur, etc.). The distorted images will influence the bounding box which is the key to detecting objects. The distortions not only blind the human eye, but also fraud detector to make the confidence of the anchor reduce.

To tackle this challenging problem, Huang, Le, and Jaw~\cite{7} employed two subnetworks to jointly learn visibility enhancement and object detection, where the impact of image degradation is reduced by sharing the feature extraction layers. However, it is hard to tune the parameters to balance the weights between detection and restoration during training. Another approach is to eliminate the effects of image distortion by pre-processing the input image (e.g., image defogging, image de-rain, image enhancement, etc.).

\begin{figure}[t]
	\centering
	\setlength{\abovecaptionskip}{0pt}
	\setlength{\belowcaptionskip}{0pt}
	\includegraphics[scale=0.8]{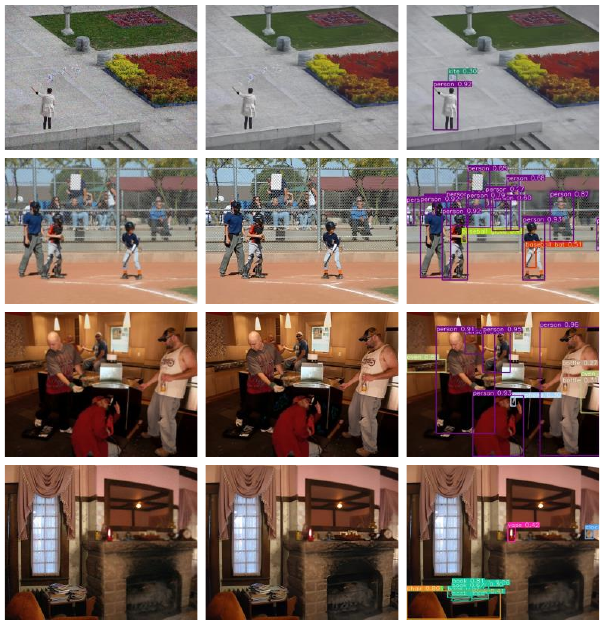}
	\caption{\label{figfront} The distorted images denoised by Restormer and detected by our model. Text on detection box represents class of CD-COCO. The number represents the confidence of predicted class.}
\end{figure}

In this paper, our approach is also constructed according to this paradigm. We apply YOLOv7~\cite{8} and ensemble model to address the problem of object detection under uncontrolled acquisition environment, which are mainstream approaches in object detection tasks. Because these models have difficulty identifying object locations and classes where the image is affected by specific environments and noise, the model is improved by adding a denoiser to the model to increase its resistance to specific noise and reduce the noisy image to a normal image to make the model better at detecting images. Considering that the dataset contains several types and intensities of noise, multiple denoisers are integrated and trained to ensure that the model can remove factors from the real scene that interfere with the captured images. Experimental results demonstrate that our model achieves excellent performance on the test set with more than 56.2\% mAP(0.5:0.95).

\section{Basic algorithm}
\label{sec:format}
The challenge focuses on real scenarios where object detection is difficult due to different acquisition conditions. To address this challenge, we increase the detection accuracy of the detector by removing specific interference information and revealing more potential information to the image adaptive detection model, where the image input to the object detector is denoised so that the noise is attenuated. The complete denoiser includes a real noise denoiser, a dynamic blur denoiser, and a super-resolution model.

\begin{figure}[t]
	\centering
	\setlength{\abovecaptionskip}{0pt}
	\setlength{\belowcaptionskip}{0pt}
	\includegraphics[scale=0.31]{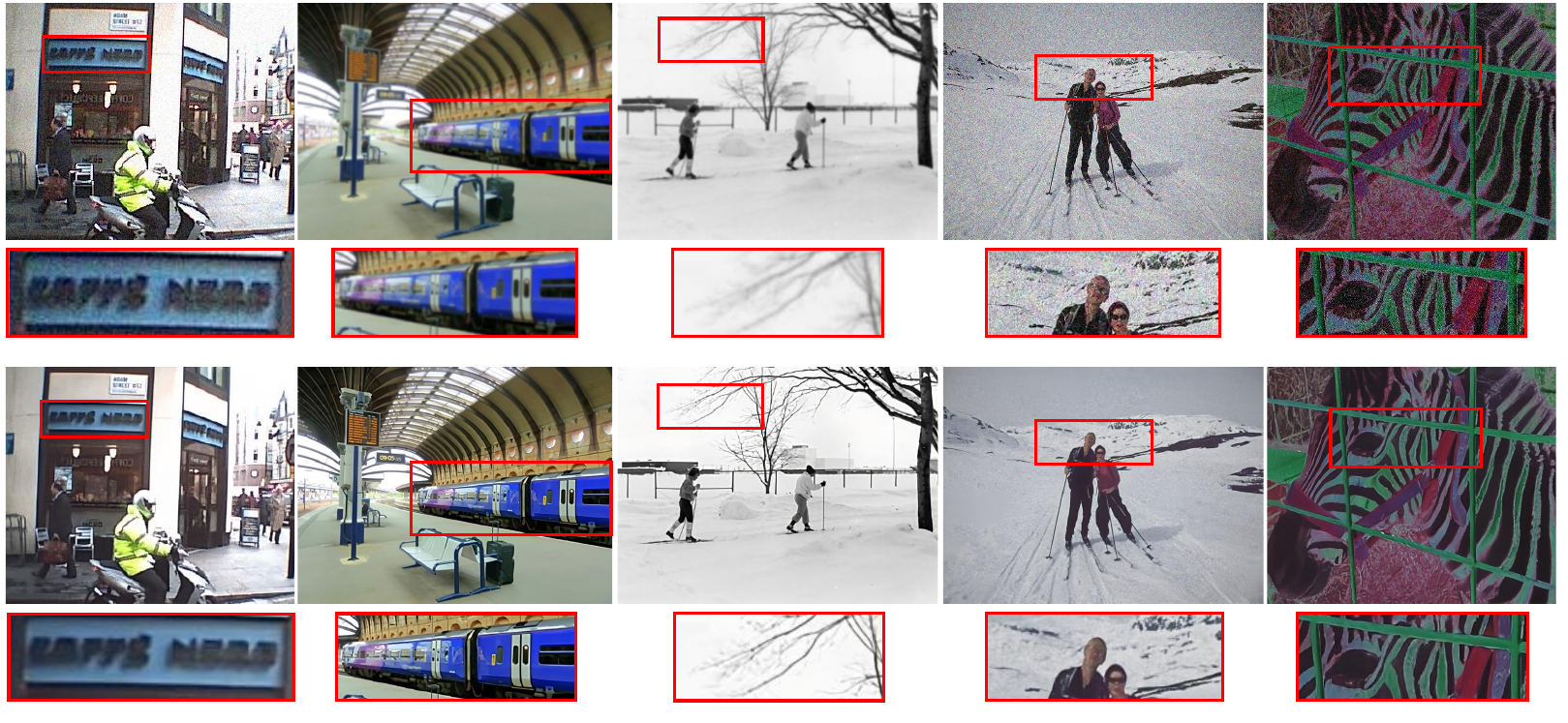}
	\caption{\label{figfront} The figure shows the performance of the purifier. The red box magnifies the detail of the image to make the denoising effect more visible. }
\end{figure}

\begin{figure}[!t]
	\centering
	\setlength{\abovecaptionskip}{0pt}
	\setlength{\belowcaptionskip}{0pt}
	\begin{subfigure}
		\centering
		\includegraphics[scale=0.37]{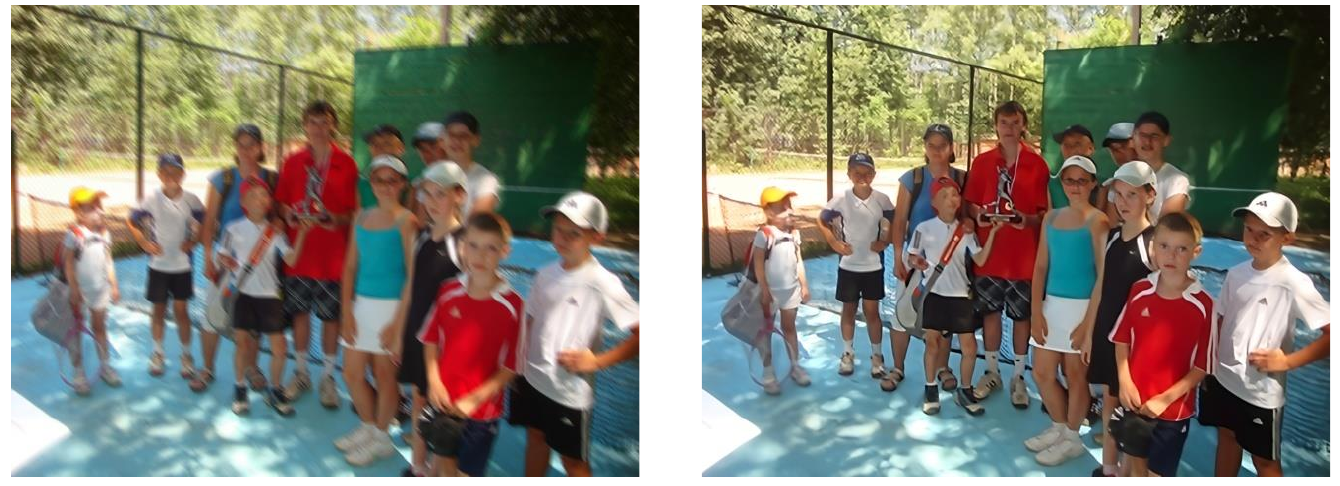}
		\label{chutian3}
	\end{subfigure}

	\begin{subfigure}
		\centering
		\includegraphics[scale=0.24]{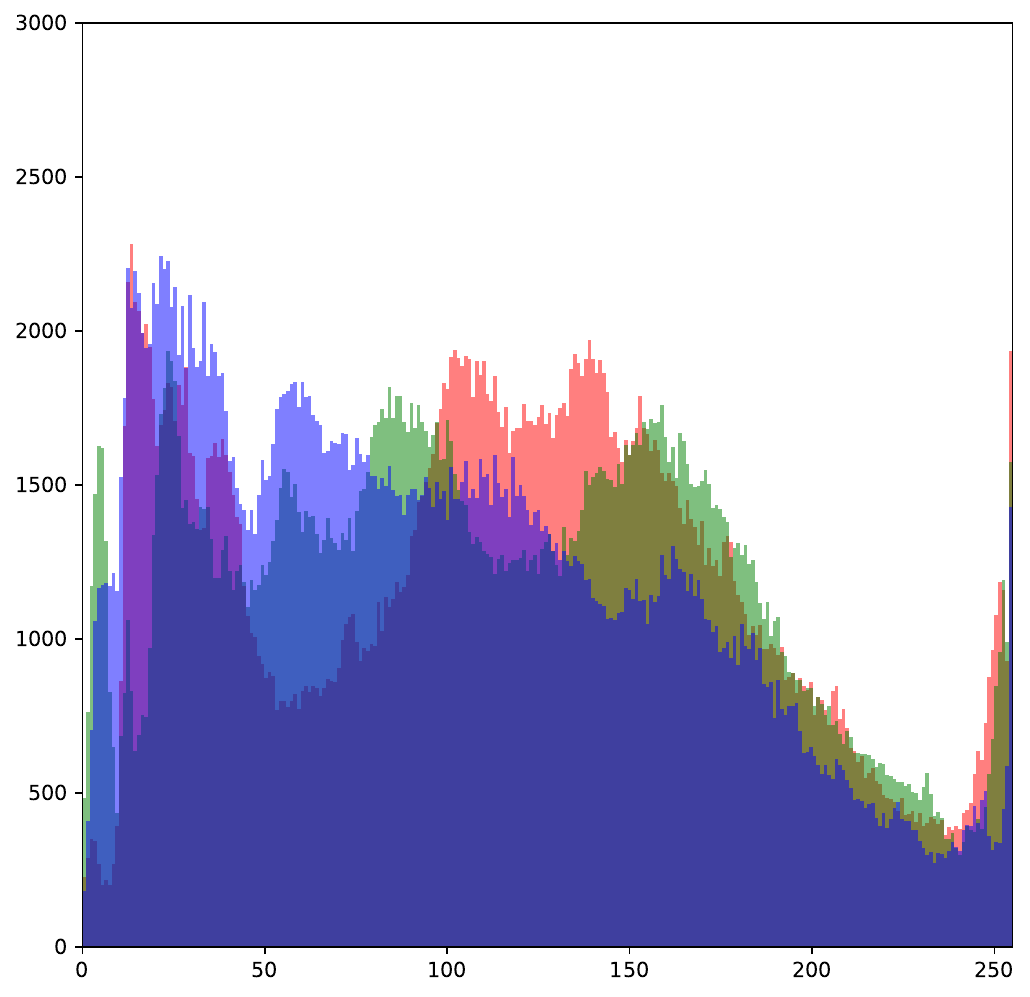}
	\end{subfigure}
	\begin{subfigure}
		\centering
		\includegraphics[scale=0.24]{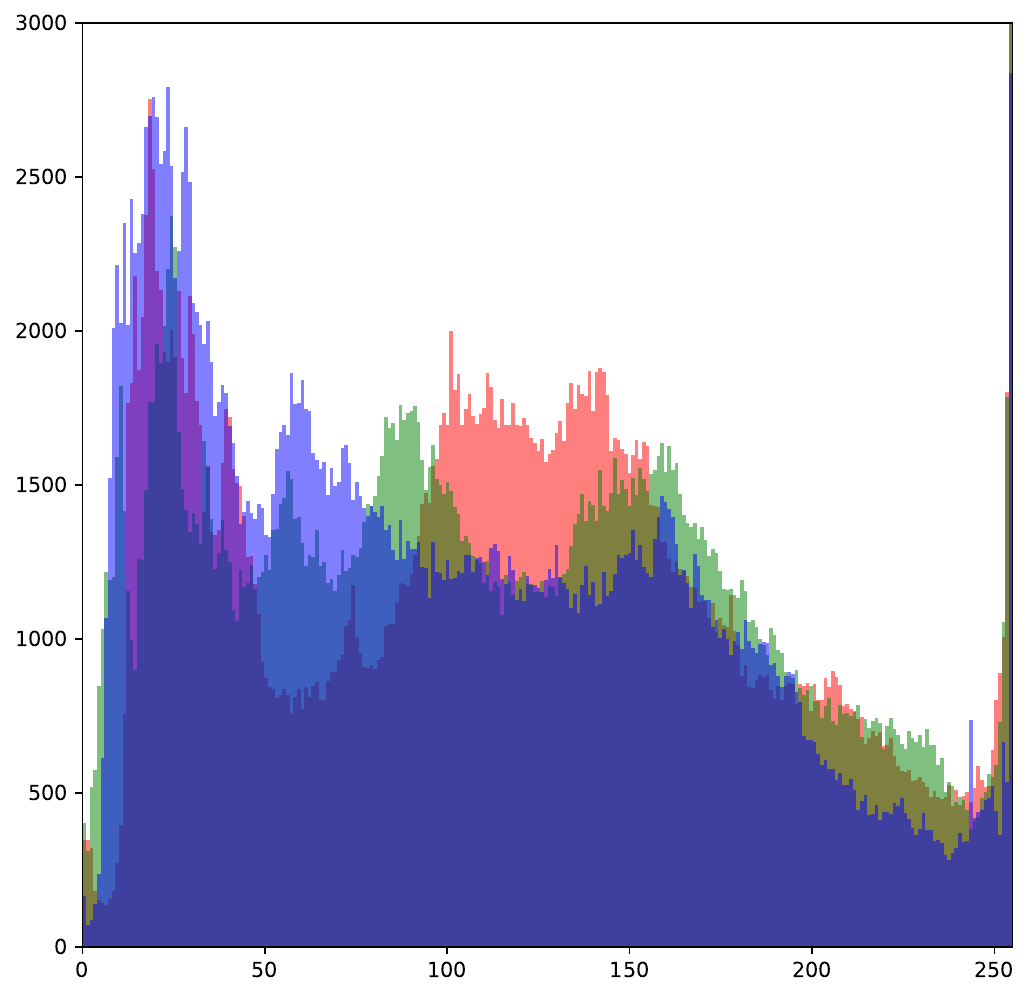}
	\end{subfigure}
	\label{fig3}
	\caption{After denoised, the pixel value distribution becomes sharper, reducing the effects of blurring. }
\end{figure}

In this paper, in terms of detectors, we used YOLOv7 as our base detection model. Real noise denoiser In general training and detection process, real noise denoiser is often used to counteract the noise that interferes with the input image of the model, such as KBNet~\cite{10}, PNGAN~\cite{11}, HINet~\cite{12}, Restormer~\cite{13}, etc. For the denoiser integration model, we chose NAFNet~\cite{9}, Restormer, and KBNet three denoisers for removing real noise.

YOLOv7 is a series of mainstream object detection algorithms developed from the paper. The largest model YOLOv7e6 makes the best performance which makes it particularly suitable for detecting images with noises.

Restormer is an image restoration transformer model that is computationally efficient to handle high- resolution images. It makes key designs for the core components of the transformer block for improved feature aggregation and transformation.

KBNet and NAFNet are networks for image restoration. KBA is the key design that adopts the learnable kernel bases to model the local image patterns and fuse kernel bases linearly to aggregate spatial information efficiently.

\begin{figure}[t]
	\centering
	\setlength{\abovecaptionskip}{0pt}
	\setlength{\belowcaptionskip}{0pt}
	\includegraphics[scale=0.37]{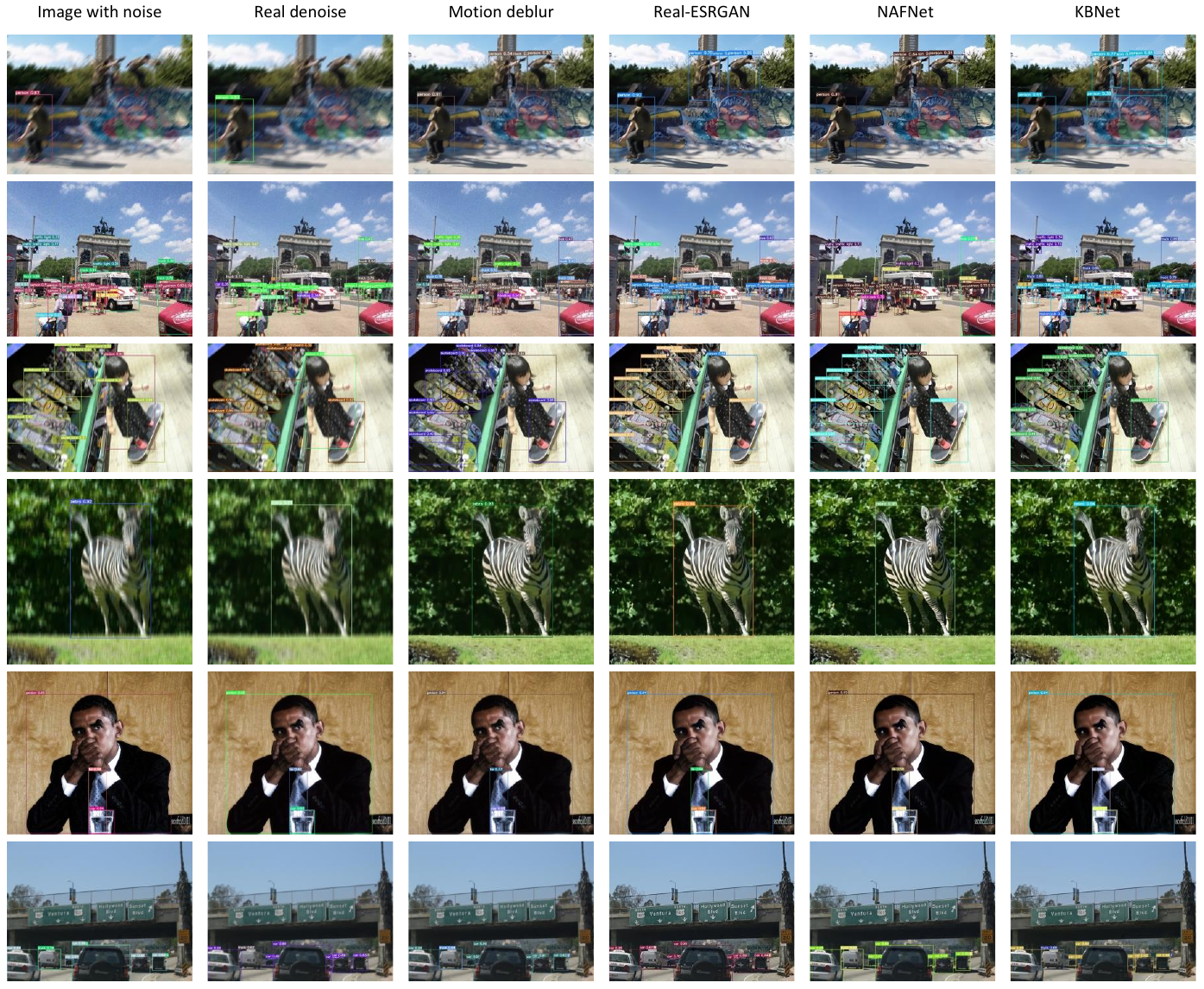}
	\caption{The figure shows the performance of different denoiser and super-resolution models. The real denoising and motion deblurring can remove the noise of the image and increase the confidence of the bounding box.}
	\label{fig4}
\end{figure}

After the detection model outputs multiple bounding boxes, we fuse the similar bounding boxes by WBF~\cite{15} to increase the confidence of bounding boxes and IOU. Fig.~\ref{fig4} shows that our method improves the performance of the detector.

\begin{figure*}[htbp]
	\centering
	\setlength{\abovecaptionskip}{0pt}
	\setlength{\belowcaptionskip}{0pt}
	\includegraphics[scale=0.65]{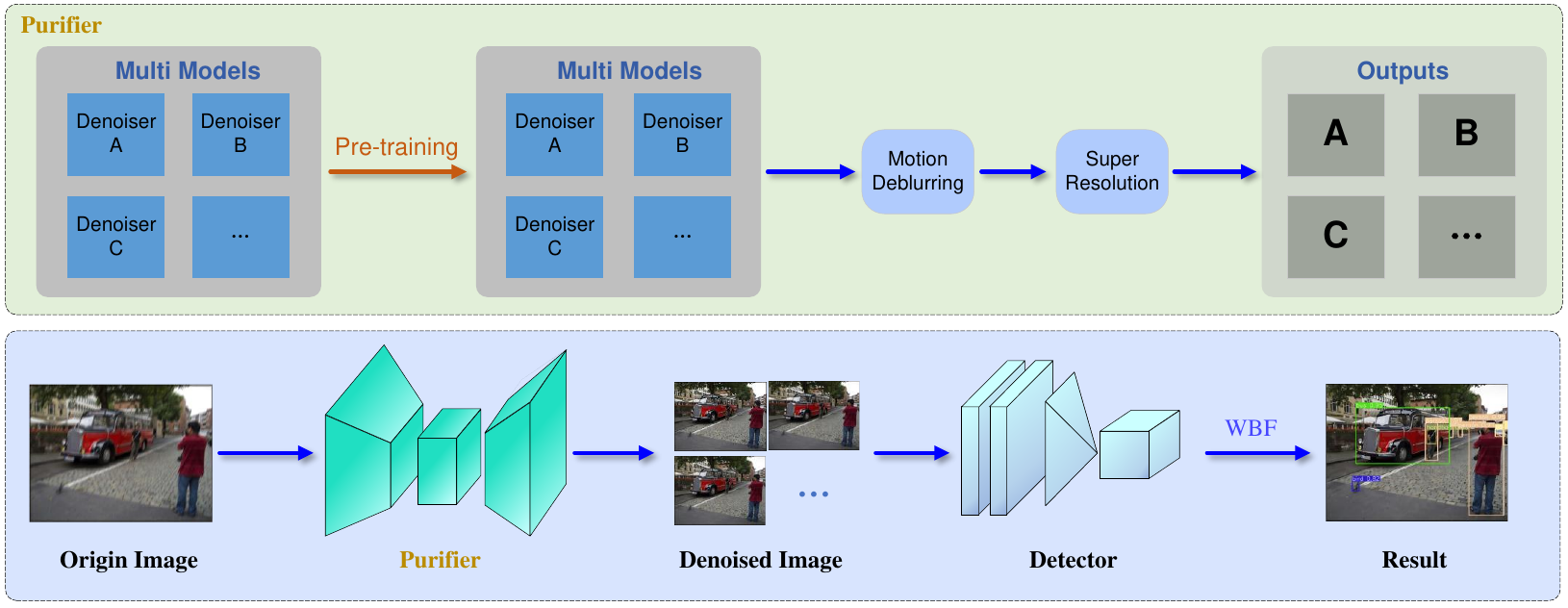}
	\caption{The figure shows the entire implementation procedure. We pretrain multiple denoise models on SIDD~\cite{16} and combine them as the purifier. The detection model is trained on augmented training dataset. We apply Weighted Boxes Fusion(WBF) to ensemble the prediction of multiple denoised images. }
	\label{fig5}
\end{figure*}

\section{Implementation}
\subsection{pre-processing}
The original dataset obtains 93,811 images of 80 classes on the training set. And the validation set has 4,926 images. The test set contains 23,527 images. They added dedicated distortions type at specific levels according to the scene context of images. It was correlated to the scene type and context. The distortion severity level will change with the object type and position for local distortions or atmospheric distortions. During the data processing, we apply the ensemble model and data augmentation. Other related parameters are showed on Tab.~\ref{table1}

\begin{table}[t]
	\centering
	\setlength{\abovecaptionskip}{0pt}
	\setlength{\belowcaptionskip}{10pt}
	\caption{Different denoiser and super-resolution model setting and special modules }
	\setlength{\tabcolsep}{0.1cm}{
		\begin{tabular}{cc}
			\toprule[1.5pt]
			Model       & settings       \\ \hline
			Restormer~\cite{13}  & \makecell{MDTA attention head = [1,2,4,8] \\ GDFN channel expansion factor = 2.66}                                                          \\
			\hline
			KBNet~\cite{10}   &  \makecell{Encoder block = \{2,2,2,2\} \\ Decoder block = \{4,2,2,2\} \\ Iteration = 300k \\ Kernel base number = 32 }                                                          \\
			\hline
			NAFNet~\cite{8}    &  \makecell{Optimizer = Adam \\ Number of GMAC = 16 \\ Iteration = 200k \\ Width and number of blocks = 32,36 \\ Input size = $256\times256$ }                                                           \\
			\hline
			Real-ESRGAN~\cite{14}    &  \makecell{Optimizer = Adam \\ Weight of feature map = \{0.1,0.1,1,1,1\} \\ Iteration = 400k \\ Input size = $256\times256$ }                                                           \\
			\bottomrule[1.5pt]
	\end{tabular}}
	\label{table1}
\end{table}

\subsection{Training Procedure}
YOLOv7: The model YOLOv7e6 has been pre-trained by the COCO dataset. Then we train the model on the distorted training set. We choose Adam as the optimizer with momentum=0.999 and weight decay is 0.0005. The learning rate is set at 0.001. It will be updated by cosine cyclical learning rate strategy. The box loss gain, classification loss gain, and object loss gain is 0.05, 0.3 and 0.7.

\subsection{Testing}
During the testing phase, we first apply the ensemble model module. The test examples are denoised by Restormer, NAFNet, and KBNet. The multiple denoised images will be achieved. Then we use a motion deblurring model to eliminate the motion blur noise in the images. Because most denoising models cause low-resolution images. The Real-esrgan~\cite{14} can restore low-resolution images with unknown and complex degradations. Lastly, we apply the super-resolution to increase visual performance and improve image quality.

From Fig.~\ref{fig3}, we can know that the images distorted by severe environments can not be detected by the human eye. The pixel value distribution is smooth when the images are distorted by motion blur. The histogram can measure the effectiveness of our denoising algorithm, as the distribution of pixel values in an image will usually change after denoising. A good denoising algorithm should be able to retain as much detail as possible in the image while reducing the effect of noise. Therefore, if the distribution of pixel values in the denoised image is more concentrated, the denoising is more effective.

\begin{table*}[t]
	\centering
	\setlength{\abovecaptionskip}{0pt}
	\setlength{\belowcaptionskip}{10pt}
	\caption{ Experimental results of YOLOv7e6 on test set. Ticked item represents that corresponding method is applied during the procedure. Aug. represents data augmentation.}
	\begin{tabular}{cccccccc|cccc}
		\toprule[2pt]
		Aug.  & \makecell{Real \\ Denoise}  & \makecell{Motion \\ Deblur} & \makecell{Real \\ ESRGAN} &\makecell{Large \\ size} & KBNet & NAFNet &  WBF & Precision & Recall & \makecell{mAP \\ (0.5)} & \makecell{mAP \\ (0.5:0.95)}  \\ \midrule
		&&&&&&&&0.785&0.635 &0.706&0.511 \\ 
		\Checkmark &&&&&&&&0.752& 0.654&0.698&0.513 \\ 
		\Checkmark &\Checkmark&&&&&&&0.745& 0.620&0.678&0.494 \\ 
		\Checkmark &\Checkmark&\Checkmark&&&&&& 0.759& 0.643&0.700&0.517 \\ 
		\Checkmark &\Checkmark&\Checkmark&\Checkmark&&&&&0.777&0.635&0.709&0.510  \\ 
		\Checkmark &\Checkmark&\Checkmark&\Checkmark&\Checkmark&&&&0.787&0.660&0.731&0.538 \\ 
		\Checkmark &\Checkmark&\Checkmark&\Checkmark&\Checkmark&\Checkmark& & &0.774&0.659&0.731&0.539 \\
		\Checkmark &\Checkmark&\Checkmark&\Checkmark&\Checkmark&\Checkmark&\Checkmark& &0.773&0.669&0.736&0.545 \\
		\Checkmark &\Checkmark&\Checkmark&\Checkmark&\Checkmark&\Checkmark&\Checkmark&\Checkmark&0.770&0.671&0.741&0.550 \\
		\bottomrule[2pt]
	\end{tabular}
	\label{table2}
\end{table*}

\begin{table}[!t]
	\centering
	\setlength{\abovecaptionskip}{0pt}
	\setlength{\belowcaptionskip}{10pt}
	\caption{Different model sequence will cause different denoised images and offset the bounding box.  }
	\setlength{\tabcolsep}{0.1cm}{
		\begin{tabular}{ccccc}
			\toprule[1.5pt]
			\makecell{Model \\ sequence}      & Precision & Recall & \makecell{mAP \\ (0.5)} & \makecell{mAP \\ (0.5:0.95)}       \\ \hline
			RD\textrightarrow RE\textrightarrow MD  &  0.776 &  0.670  & 0.738 &   0.546                                 \\
			\hline
			MD\textrightarrow RD\textrightarrow RE   &    0.787 & 0.662& 0.737&    0.546                                             \\
			\hline
			MD\textrightarrow RE\textrightarrow RD    & 0.772   & 0.662&0.731 &  0.538                                            \\
			\hline
			RD\textrightarrow MD\textrightarrow RE    & 0.770&0.671&0.741&0.550                                                           \\
			\bottomrule[1.5pt]
	\end{tabular}}
	\label{table3}
\end{table}

\subsection{Model Ensemble}
Not only in the denoiser we use the ensemble model but also in the bounding box selection of detector we use WBF as our ensemble strategy. Different from the common model, we put multiple denoised images into the detector, generating a huge number of bounding boxes. 

WBF method uses confidence scores of all proposed bounding boxes in the iterative algorithm that constructs the averaged boxes. After ensemble, our detector can predict a more accurate anchor on the object with higher confidence. The weights of the multiple images will be related to the performance of the corresponding denoiser on the SIDD. The entire procedure of our implementation is illustrated in Fig.~\ref{fig5}.

\section{Ablation studies}
\label{sec:majhead}
The main results of the detection models YOLOv7e6 on the test set are summarized in Tab.~\ref{table2}. The components of the purifier are the most important module in our method. Most denoiser are benefit to the model's performance. Among them, the super-resolution increases the mAP of dataset, performing effectively. It suggests that the ensemble model is necessary for the detection task in the severe environment.

In Tab.~\ref{table3}, it shows that the model sequence also affects image rebuild and the performance of the detector. The RD represents the real denoiser. The MD represents the motion deblur model. The RE represents the Real-ESRGAN. The real denoiser and motion deblur model will make disturbed images low resolution, so if the Real-ESRGAN is the first model it can not make images high resolution. It will decrease the confidence of the bounding box of the detector.

\section{Final result}
\label{sec:print}
Final result of our ensemble denoiser and detection model on ICIP 2023 challange of Object detection under uncontrolled acquisition environment and scene context constraints is presented on Tab.~\ref{table4}.

\begin{table}[!t]
	\centering
	\setlength{\abovecaptionskip}{0pt}
	\setlength{\belowcaptionskip}{10pt}
	\caption{The final result of YOLOv7e6 }
	\setlength{\tabcolsep}{0.1cm}{
		\begin{tabular}{ccccc}
			\toprule[1.5pt]
			Model     & Precision & Recall & \makecell{mAP \\ (0.5)} & \makecell{mAP \\ (0.5:0.95)}       \\ \hline
			YOLOv7e6  & 0.770 & 0.682 & 0.751 & 0.562                                         \\
			\bottomrule[1.5pt]
	\end{tabular}}
	\label{table4}
\end{table}

\section{Conclusion}
We investigate an integrated denoising object detector for challenge dataset containing multiple noises. The denoiser is an important component in our detection process. We form an integrated denoising model by fusing multiple denoisers trained by WBF. With the integrated denoising model, the challenge dataset containing multiple noises is not only effectively removed from multiple noises, but also the distribution of its pixel values becomes clearer. Our detector is based on YOLOv7e6e, and the model is optimized by integration training, which enables the detection model to predict objections with higher confidence. Our model achieves excellent performance on the Challenge dataset with an mAP(0.5:0.95) of over 56.2\%.



\bibliographystyle{IEEEbib}
\bibliography{wenxian.bib}

\end{document}